\def\year{2021}\relax
\documentclass[letterpaper]{article} % DO NOT CHANGE THIS
\usepackage{aaai21}  % DO NOT CHANGE THIS
\usepackage{times}  % DO NOT CHANGE THIS
\usepackage{helvet} % DO NOT CHANGE THIS
\usepackage{courier}  % DO NOT CHANGE THIS
\usepackage[hyphens]{url}  % DO NOT CHANGE THIS
\usepackage{graphicx} % DO NOT CHANGE THIS
\usepackage{natbib}  % DO NOT CHANGE THIS AND DO NOT ADD ANY OPTIONS TO IT
\usepackage{caption} % DO NOT CHANGE THIS AND DO NOT ADD ANY OPTIONS TO IT

\usepackage{amsmath}
\usepackage{amssymb}
\usepackage{mathtools}
\usepackage{booktabs}
\usepackage[switch]{lineno} % line numbers
\usepackage{multirow}
\usepackage{listings}

\nocopyright

\title{Robustness to Missing Features using Hierarchical Clustering\\ with Split Neural Networks}
\author{
    Rishab Khincha\textsuperscript{\rm 1, 2},
    Utkarsh Sarawgi\textsuperscript{\rm 1},
    Wazeer Zulfikar\textsuperscript{\rm 1},
    Pattie Maes\textsuperscript{\rm 1}
    \\
}
\affiliations{
    \textsuperscript{\rm 1}Massachusetts Institute of Technology, USA \\
    \textsuperscript{\rm 2}BITS Pilani Goa Campus, India \\
    \{rkhincha, utkarshs, wazeer, pattie\}@mit.edu

}
\begin{document}
% \linenumbers

\maketitle

\begin{abstract}
The problem of missing data has been persistent for a long time and poses a major obstacle in machine learning and statistical data analysis. Past works in this field have tried using various data imputation techniques to fill in the missing data, or training neural networks (NNs) with the missing data. In this work, we propose a simple yet effective approach that clusters similar input features together using hierarchical clustering and then trains proportionately split neural networks with a joint loss. We evaluate this approach on a series of benchmark datasets and show promising improvements even with simple imputation techniques. We attribute this to learning through clusters of similar features in our model architecture. The source code is available at \lstinline|https://github.com/usarawgi911/Robustness-to-Missing-Features|
\end{abstract}

\section{Introduction}
Learning in the regime of incomplete or missing data has been a fundamental problem in machine learning. It presents various limitations - it reduces the statistical power of the data, induces a bias when estimating parameters and is not a good representation of the original underlying distribution. With the increasing use of neural networks in various domains, it is important to build techniques that can easily extend and improve the current algorithms. This can also help improve performance in case of missing data at test time.

% The pattern of missing values in data is broadly classified into 3 categories - (a) missing completely at random or MCAR when there is no relation between the missing instances and any other data feature; (b) missing at random (MAR) when the missing instances are only affected by the observable features; and (c) missing not at random (MNAR) when the data instance missing depends on the feature itself.

Various statistical imputation strategies (for eg. mean, k-nn imputation) have been suggested to fill the missing attributes based on the observed data \cite{mcknight2007book, murray2018multiple}. One can also learn separate models like NNs to train on the observed data and predict the missing values \cite{sharpe1995nn}. Recently, methods have been suggested to train NNs directly with missing data without any imputations \cite{nips2018processing}.

\begin{figure}[h]
\centering
\includegraphics[width=\linewidth]{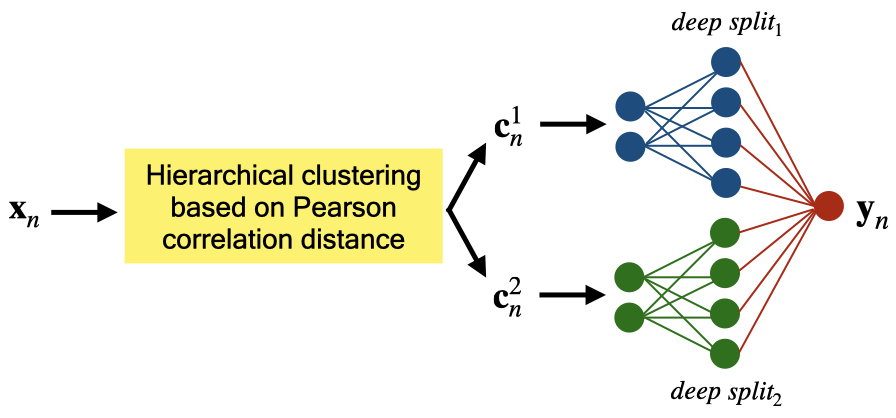}
\caption{Process diagram of our method - feature clustering followed by a Split NN}
\label{splitnn}
\end{figure}

We propose a simple procedure that first clusters similar or statistically correlated input features together, and then trains proportionately `split neural networks' (split NNs) with these input clusters using a joint loss (Figure \ref{splitnn}). We show that learning through these clusters of similar features in our model architecture achieves results comparable to other suggested methods in the literature. Our method is effective and can be readily applied to most NN based architectures with just minimal changes in their data pipeline.    

% The missing attributed can be filled using any of the popular imputation strategies like mean, k-nn imputation, multiple imputations or learnt using neural network approaches.   
% - introduce the problem : learning from incomplete data is hard, give example of sensors etc. Reduces statistical power of the data, induces bias and does not represent the data well
% - types of missing data : mcar, mar, mnar  
% - related work : mostly data imputation strategies, talk about them. NIPS 2018 paper on processing with missing features, and relevant related works from there
% - our solution overview : split features based on hc clustering, or domain needs. highlight why it works

\begin{table}[hb!]
% \small
\centering
\begin{tabular}{lcccc}
\toprule
Dataset & Samples & Features & Missing & $k$ \footnotemark\\ \midrule
bands   & 539     & 19       & 5.38\%  & 10\\
kidney disease  & 400     & 24       & 10.54\%  & 9 \\
hepatitis   & 155     & 19       & 5.67\%  & 14  \\
horse   & 368     & 22       & 23.80\%  & 14 \\
mammographics   & 961     & 5        & 3.37\%  & 4  \\
pima    & 768     & 8        & 12.24\%  & 7 \\
winconsin   & 699     & 9        & 0.25\%   & 6 \\
life expectancy & 2938    & 21       & 43.7\%  & 8   \\
\bottomrule
\end{tabular}
\caption{Dataset details}
\label{dataset_table}
\end{table}
% \footnotetext{{$k$ denotes the number of feature clusters formed when using the threshold of $0.5*Pearson Correlation Distance_{max}$ during hierarchial clustering}}
\begin{table*}
% \small
\centering
\begin{tabular}{lccccccc}
\toprule
Dataset & karma & mice & mean & dropout & \citeauthor{nips2018processing} & Vanilla NN & Split NN (ours) \\ 
\midrule
bands & 0.580 & 0.544 & 0.545 & 0.616 & 0.598 & 0.551 $\pm$ 0.058 & \textbf{0.662 $\pm$ 0.051} \\
kidney disease & \textbf{0.995} & 0.992 & 0.985 & 0.983 & 0.993 & 0.972 $\pm$ 0.030 & 0.963 $\pm$ 0.032 \\
hepatitis & 0.665 & 0.792 & 0.825 & 0.780 & 0.846 & 0.716 $\pm$ 0.069 & \textbf{0.849 $\pm$ 0.075} \\
horse & 0.826 & 0.820 & 0.793 & 0.823 & \textbf{0.864} & 0.794 $\pm$ 0.036 & 0.826 $\pm$ 0.020 \\
mammographics & 0.773 & 0.825 & 0.819 & 0.814 & \textbf{0.831} & 0.827 $\pm$ 0.026 & \textbf{0.829 $\pm$ 0.016} \\
pima & 0.768 & 0.769 & 0.760 & 0.754 & 0.747 & 0.762 $\pm$ 0.020 & \textbf{0.777 $\pm$ 0.039} \\
winconsin & 0.958 & \textbf{0.970} & 0.965 & 0.964 & \textbf{0.970} & 0.961 $\pm$ 0.015 & 0.964 $\pm$ 0.009 \\
\bottomrule

\end{tabular}
\caption{Classification accuracies on benchmark datasets (other methods do not report the performance variance across folds)}
\label{uci_table}
\end{table*}

\section{Process architecture}
\textbf{Notation and setup:} Let $\mathbf{x} \in \mathbb{R}^d$ represent a set of $d$-dimensional input features and $y \in \mathbb{R}$ denote the real-valued label for classification or regression.  Given a training dataset $\mathcal{D} = \{(\mathbf{x}_n, y_n)\}_{n=1}^N$ consisting of N i.i.d. samples, we use a neural network with parameters $\theta$ to model the probabilistic predictive distribution $p_\theta (y|\mathbf{x})$. We split the $d$ input features of $\mathbf{x}$ into $k$ exhaustive clusters, $k > 1$, each containing $m_i$ number of features, where a feature can belong to only 1 cluster. The $i^{th}$ feature cluster containing $m_i$ input features of the $n^{th}$ data point $\mathbf{x}_n$ is denoted by $\mathbf{c}^i_n \in \mathbb{R}^{m_i}$. Thus, $\{(\mathbf{c}^i_n, y_n)\}_{n=1}^N$ represents  the $i^{th}$ input feature cluster and corresponding label for each of the N samples. Note that the label $y_n$ is the same across all the input clusters $c^i_n$ corresponding to the $n^{th}$ data point.

\textbf{Feature clustering:} The input feature space is split into $k$ exhaustive clusters using  hierarchical clustering based on Pearson correlation distance. The dendograms thus obtained upon hierarchical clustering with complete linkage are thresholded relative to the maximum distance to obtain feature clusters. (we chose 50\% for Table \ref{dataset_table}, and can be changed to control the number of clusters $k$) Note that we are clustering features, which should not be confused with clustering datapoints. Splitting the input features in this way is effective since the cluster of similar features tend to work well together to substitute for the missingness, to provide better estimates.

\textbf{Split NN:} The NN is then split with hidden units in each of the deep splits proportional to the number of features in the corresponding clusters, as shown in Figure \ref{splitnn}. The split NN is then trained with all feature clusters using a joint loss (categorical cross-entropy for classification and mean-squared-error for regression), wherein the missing values are imputed for the mean value of that input feature.

% - notation, set up : basic math formulation
% - algorithm : discuss hc clustering, training procedure

\section{Experiments and Results}
We evaluate our approach on a series of benchmark datasets for classification tasks (details in Table \ref{dataset_table}) used by \citeauthor{nips2018processing} to allow for fair comparisons. Our network consists of 50 hidden units with ReLU activations, trained to optimize for the categorical cross-entropy loss. We use a 5-fold double cross-validation setup to report classification accuracies and train all the networks with a learning rate of 0.01 and a batch size of 100 for 1000 epochs. Table \ref{uci_table} shows that our method achieves results competitive with other state-of-the-art methods (including karma, mice, mean, and dropout) as reported by \citeauthor{nips2018processing}. We also compare our performance with a vanilla NN using the same model architecture as ours without any feature clustering and NN splitting.  

In real-world settings, it is apparent that some features might be absent at inference time. While the performance of NNs usually drop, we observe that Split NNs are relatively robust to it as a consequence of statistically correlated features clustered together. We demonstrate this with a regression dataset `Life Expectancy (WHO)' (details in \ref{dataset_table}). We train a vanilla NN as well as a split NN, each with a hidden layer of 50 units as before. We test our network on all the data points containing any missing values (43.7\% of the full dataset) and split the remaining dataset into a 80-20 train-val split. We repeat our procedure for $k=8$ and $2$ clusters, and observe that Split NN achieves test results better than that than a vanilla NN (Table \ref{life_table}).

% Talk about datasets, relation to the NIPS 2018 paper
% Life expectancy dataset too

\begin{table}[ht!]
\centering
  \centering
  \begin{tabular}{lcc}
    \toprule
    Model & Val RMSE & Test RMSE \\
    \midrule
    Vanilla NN & 3.882 & 5.116 \\
Split NN ($k=8$) & 2.945 & 4.246 \\
Split NN ($k=2$) & 3.584 & \textbf{4.006} \\
    \bottomrule
  \end{tabular}
  \caption{RMSE scores on the Life Expectancy dataset}
  \label{life_table}
\end{table}
% \footnotetext{$k=2$ feature clusters is when when using the threshold of $0.75*Pearson Correlation Distance_{max}$ during hierarchical clustering}

\section{Conclusion}
We have proposed a conceptually simple yet effective change to neural network architectures to produce more robust predictions in case of missing data. Creating clusters of statistically correlated input features show impressive performance even with using simple imputation techniques. Learning and inferring from data with incomplete features has been a pervasive problem in machine learning and statistical analysis. Various real-life applications in medical data, sensor data and pilot studies suffer due to the loss in performance and robustness due to missing data. We are very excited with the initial results and the future avenues this work opens up.

\bibliography{main}

\begin{thebibliography}{4}
\providecommand{\natexlab}[1]{#1}
\providecommand{\url}[1]{\texttt{#1}}
\providecommand{\urlprefix}{URL }
\expandafter\ifx\csname urlstyle\endcsname\relax
  \providecommand{\doi}[1]{doi:\discretionary{}{}{}#1}\else
  \providecommand{\doi}{doi:\discretionary{}{}{}\begingroup
  \urlstyle{rm}\Url}\fi

\bibitem[{Mcknight et~al.(2007)Mcknight, Mcknight, Sidani, and
  Figueredo}]{mcknight2007book}
Mcknight, P.; Mcknight, K.; Sidani, S.; and Figueredo, A. 2007.
\newblock \emph{Missing Data: A Gentle Introduction}.

\bibitem[{Murray(2018)}]{murray2018multiple}
Murray, J. 2018.
\newblock Multiple Imputation: A Review of Practical and Theoretical Findings.
\newblock \emph{Statistical Science} 33.
\newblock \doi{10.1214/18-STS644}.

\bibitem[{Sharpe and Solly(1995)}]{sharpe1995nn}
Sharpe, P.~K.; and Solly, R.~J. 1995.
\newblock Dealing with Missing Values in Neural Network-Based Diagnostic
  Systems.
\newblock \emph{Neural Comput. Appl.} 3(2): 73--77.
\newblock \doi{10.1007/BF01421959}.
\newblock \urlprefix\url{https://doi.org/10.1007/BF01421959}.

\bibitem[{Smieja et~al.(2019)Smieja, Łukasz Struski, Tabor, Zieliński, and
  Spurek}]{nips2018processing}
Smieja, M.; Łukasz Struski; Tabor, J.; Zieliński, B.; and Spurek, P. 2019.
\newblock Processing of missing data by neural networks.
\newblock In \emph{Advances in Neural Information Processing Systems 31}.
\newblock
  \urlprefix\url{http://papers.nips.cc/paper/7537-processing-of-missing-data-by-neural-networks.pdf}.

\end{thebibliography}
\end{document}